\documentclass{article} 
\usepackage{sty/arxiv}
\usepackage{amsmath}
\usepackage{array}
\usepackage{url}
\usepackage{booktabs}
\usepackage{multirow}
\usepackage{import}
\usepackage{amssymb}
\usepackage{mathtools}
\usepackage[table]{xcolor}
\usepackage{hyperref}
\usepackage{subfig}
\usepackage{xcolor}   
\usepackage{float}
\usepackage{enumitem}
\usepackage[flushleft]{threeparttable}
\usepackage{lipsum}

\begin{document}

\title{Temporal Persistence and Intercorrelation of Embeddings Learned by an End-to-End Deep Learning Eye Movement-driven Biometrics Pipeline}
\author{Mehedi~Hasan~Raju,
        Lee~Friedman,
        Dillon~Lohr,
        and~Oleg~Komogortsev%
\thanks{Department of Computer Science, Texas State University, San Marcos, TX, 78666 USA (e-mail: m\_r1143@txstate.edu; l\_f96@txstate.edu; djl70@txstate.edu; ok11@txstate.edu).} 
\thanks{Manuscript received - -, -; revised - -, -.}}

\date{}

%
%

\markboth{IEEE Transactions on Biometrics, Behavior, and Identity Science
,~Vol.~x, No.~y, Month~Year}%
{Raju \MakeLowercase{\textit{et al.}}}
%



\maketitle
\begin{abstract}
What qualities make a feature useful for biometric performance?  
In prior research, pre-dating the advent of deep learning (DL) approaches to biometric analysis, a strong relationship between temporal persistence, as indexed by the intraclass correlation coefficient (ICC), and biometric performance (Equal Error Rate, EER) was noted.  
More generally, the claim was made that good biometric performance resulted from a relatively large set of weakly intercorrelated features with high ICC.  
The present study aimed to determine whether the same relationships are found in a state-of-the-art DL-based eye movement biometric system (``Eye-Know-You-Too''), as applied to two publicly available eye movement datasets.  
To this end, we manipulate various aspects of eye-tracking signal quality, which produces variation in biometric performance, and relate that performance to the temporal persistence and intercorrelation of the resulting embeddings.  
Data quality indices were related to EER with either linear or logarithmic fits, and the resulting model $R^2$ was noted.  
As a general matter, we found that temporal persistence was an important predictor of DL-based biometric performance, and also that DL-learned embeddings were generally weakly intercorrelated. 
\end{abstract}


%

\section{Introduction}

Biometric systems are increasingly replacing traditional (i.e., password-based, PIN-based, etc) authentication systems for security due to their reliability and convenience \cite{alrawili2023comprehensive}.  
According to Nelson \cite{Nelson}, biometric features should be ``...reliable, unique, collectible, convenient, long term, universal, and acceptable''.   
Features with relative permanence are required for good biometric performance \cite{van2013characteristics,harvey2017permanence,das2022advancing,bargary2017individual}.  
Friedman et al. \cite{friedman2017method} was the first to evaluate the intraclass correlation coefficient (ICC) to index the reliability (or ``temporal persistence'') of biometric features.  
Temporal persistence ensures that the chosen biometric features remain relatively stable over the relevant period, enabling consistent and accurate authentication.
In the era before the wide advent of deep-learning (DL) based biometric systems, Friedman et al. found that more temporally persistent features lead to better performance in biometric systems i.e., feature sets with high ICC biometrically outperformed features with comparatively low ICC.  
The extent to which the ICC is useful for DL-based biometric embeddings is unknown.  
The main goal of the present study is to evaluate this question.  
If it is found that temporal persistence is important for the biometric performance of DL-based systems, this would tend to generalize the importance of this quality to all biometric systems.  

Friedman et al \cite{Temporal} assessed why temporal persistence affects biometric performance.  Recall that the computation of the EER requires the creation of a distribution of genuine and impostor similarity scores.  Friedman et al. found that the median of the genuine scores distribution increases and the spread (interquartile range) decreases with increasing ICC.  However, the impostor distributions do not change.  These changes in the genuine similarity score distributions lead to a better separation from the impostor distributions and therefore a lower EER. 

Physiological and anatomical biometric systems, like fingerprint and facial recognition, rely on physical characteristics that can change over time due to aging or injury. 
This can downgrade biometric performance. 
To address this limitation, researchers have been exploring behavioral biometrics that are more likely to remain stable (e.g., voice, gait, signature recognition). 
One such approach is eye movement biometrics, which has emerged as a promising behavioral biometric modality, attracting significant attention\cite{kasprowski2004, katsini2020role, mh2023determining, mh2023filtering}.
Unique and consistent patterns of eye movement offer advantages like user identification \cite{schroder2020robustness, rigas2017current, deepeyedentification, deepeyedentificationlive}, user authentication \cite{lohr2020metric, Lohr2020, lohrTBIOM,lohr2022eye, raju2024signal}, high specificity\cite{bargary2017individual}, disorder detection \cite{nilsson2016screening, billeci2017integrated}, gender prediction \cite{sargezeh2019gender, al2020gender},  resistance to impersonation \cite{eberz2015preventing, Komogortsev2015} and liveness detection \cite{rigas2015, raju2022iris}. 
Building on the foundation of temporal persistence in the traditional biometric approach, our study shifts the focus to DL-based behavioral biometric systems, particularly those that analyze eye movements.

This research aims to assess the role of temporal persistence and embedding's intercorrelation in a DL-based eye-movement biometric system.  
To this end, various aspects of eye-movement signal quality were manipulated to produce variations in biometric performance (EER).  
The relationship between the temporal persistence of DL-based embeddings and EER was assessed under several conditions.  
Also, the intercorrelation of sets of embeddings is evaluated.
In this paper, we will try to address the following research questions:

\begin{enumerate}[label=RQ\arabic*.]

\item  \textbf{Do reductions in the sampling rate affect biometric performance, and are these changes related to the reliability of the learned embeddings?}
We will employ decimation to achieve the desired sampling level, which reduces both the sampling rate and the number of samples. Comparing the effects of sample rate reduction to only the reduction in the number of samples will help us assess separately the effects of sample rate reduction and reduced data length. This consideration leads to RQ2.

\item \textbf{Does reduced data length at a fixed sampling rate affect the reliability of the learned embeddings?}
Decimation reduces both data length and sampling rate. Here, we will investigate the effect of decreasing data length while maintaining a fixed sampling rate.

\item \textbf{Does the number of sequences affect the reliability of the learned embeddings?} 
Computational limitations require we derive embeddings for 5-seconds of data at a time.  
These 5-second intervals are referred to as ``sequences''.  
Our baseline analyses involve averaging over either 12 or 9 sets of sequences. We wanted to evaluate the analysis results based on a range of sequences.

\item \textbf{Does the quality of the eye-movement signal affect the reliability of the learned embeddings?}
We will explore how degraded spatial precision of the eye-movement signal influences the embeddings.      

\item \textbf{Does any eye-tracking data manipulation affect the intercorrelation of the learned embeddings?}
We will explore how data manipulation of various kinds affects the absolute value of the intercorrelation of the embeddings.

\end{enumerate}

This paper provides a review of the relevant literature in Section II. 
Our methodology is detailed in Section III. 
The design of our experiments is outlined in Section IV. 
Section V presents the results obtained from these experiments. 
Analysis of these results and key insights are discussed in Section VI. 
The paper concludes with final remarks in Section VII.

\section{Prior Work}

\subsection{Prior work on Temporal persistence/Biometric permanence}

In the biometric authentication field, it is widely accepted that human traits with high temporal persistence, encompassing temporal stability and permanence, are fundamental.\cite{van2013characteristics,harvey2017permanence,das2022advancing,friedman2017method,Temporal,bargary2017individual}.
Some studies focused on evaluating the biometric permanence of a modality, comparing the biometric performance of the system at different times\cite{maiorana2015permanence,harvey2017permanence,labati2013ecg,zhang2021specificity}. 
As per our knowledge, there are relatively few studies (discussed below) that have explored the relationship between the temporal persistence of individual features and biometric performance and proposed an index to measure the temporal persistence of features.

Prior research \cite{friedman2017method} introduced the use of the intraclass correlation coefficient (ICC) to the biometric community as an index of temporal persistence, although ICC has long been used as a measure of feature reliability in various fields \cite{reliability1,reliability2,reliability3}.
It is a measure of how stable a feature is over time.
Features with high ICC are more stable than those with low ICC.
The authors argued that using features with high ICC leads to better biometric performance.
They tested this on 14 datasets (8 of them were eye-movements-related data) and found that using features with high ICC resulted in superior biometric performance in most cases.
Friedman et al. \cite{Temporal} demonstrated that increased temporal persistence makes biometric systems work better for gait and eye movement datasets. 
The median of the genuine scores distribution increases with increasing ICC, and the interquartile range (IQR) narrows which means that the genuine scores become more concentrated around a higher value as ICC increases.
The median of the imposter scores distribution does not change significantly with increasing ICC meaning that the imposter scores remain spread out across a similar range of values regardless of ICC.
These changes in the distributions lead to better separation between genuine and impostor scores, which makes it easier for a biometric system to correctly classify a sample. 
The Equal Error Rate (EER), which is the point where the false acceptance rate (FAR) and false rejection rate (FRR) are equal, is also lower for higher ICC values. 
This indicates that the system is less likely to make errors (accepting an imposter or rejecting a genuine user) when the temporal persistence is higher.

\subsection{Prior work on Eye Movement Biometrics}

Kasprowski and Ober's introduction of eye movement as a biometric modality for human authentication \cite{kasprowski2004} marked a significant milestone. 
This spurred extensive research in the field of eye movement biometrics \cite{brasil2020eye, zhang2015survey, zhang2016biometrics}, primarily aimed at developing a state-of-the-art (SOTA) approach for eye movement-based user authentication. 
Two primary approaches have emerged in this domain: the statistical feature-based approach and the machine learning-based approach.
In the statistical feature-based approach, a standardized processing sequence is employed. 
It involves segmenting recordings into distinct eye movement events using event classification algorithms, followed by the creation of a biometric template comprising a vector of discrete features from each event \cite{Rigas2017}. 
However, the challenge lies in the classification of events, which can vary in effectiveness depending on the classification algorithm used \cite{Andersson2017}. Various algorithms for classifying eye-movement events have been suggested \cite{ONH, classification1, classification2, Friedman2018}, aiming to enhance biometric performance. 
Several studies have utilized this approach, including \cite{Lohr2020, friedman2017method, Li2018, rigas2012biometric, rigas2016biometric, holland2013complex}.
Meanwhile, there's been a significant increase in the application of end-to-end machine learning approaches in eye movement biometrics. 
Recent studies have focused on deep learning, adopting two main approaches: processing pre-extracted features as in \cite{george2016score, lohr2020metric}, and learning embeddings directly from the raw eye tracking signals \cite{Jia2018, lohrTBIOM, deepeyedentification, deepeyedentificationlive, abdelwahab2022deep}.
The development of the Eye Know You Too (EKYT) \cite{lohr2022eye} model by Lohr et al. represents a significant advancement.
As per our knowledge, EKYT is a state-of-the-art (SOTA) user authentication system based on eye movement data. 
EKYT is developed in such a way that it is capable of learning meaningful embeddings. 
Embeddings offer a way for deep learning models to represent complex data in a simplified, lower-dimensional space, preserving inherent patterns and relationships.
This approach allows the model to group similar data points closer together in a vector space, facilitating the discovery of underlying patterns that might be challenging for humans to identify directly. 
Unlike traditional feature extraction, embeddings enable the model to learn these representations, potentially unveiling complex relationships within the data that enhance authentication processes. 
Once learned, these embeddings can be used in classification problems such as the authors did for eye-movement-based biometric authentication in \cite{lohr2022eye}.

Concluding our review, it's important to note that, based on our current understanding, there has been no investigation into the analysis of embeddings derived from a deep learning model in an EMB-driven pipeline. This paper will specifically address and explore this area.

\section{Methodology}

\subsection{Dataset}
We have employed two large datasets in our study. 
One was collected with a high-end eye-tracker and the other was collected with an eye-tracking-enabled virtual reality (VR) headset. 
The reason behind using two datasets is to ensure the generalizability of the findings.

The first dataset, we used in this study is the publicly available GazeBase(GB) dataset \cite{griffith2021gazebase}. 
Eye movement recordings of this dataset are collected with a high-end eye-tracker, EyeLink 1000 at a sampling rate of 1000~Hz.
It includes 12,334 monocular recordings (left eye only) from 322 college-aged subjects.
The data was collected over three years in nine rounds (Round 1 to Round 9). 
Each recording captures both horizontal and vertical movements of the left eye in degrees of visual angle.
Participants completed seven eye movement tasks: random saccades (RAN), reading (TEX), fixation (FXS), horizontal saccades (HSS), two video viewing tasks (VD1 and VD2), and a video-gaming task (Balura game, BLG). 
Each round comprised two recording sessions of the same tasks per subject, spaced by 20 minutes.
Further details about the dataset and recording procedure are available in \cite{griffith2021gazebase}.

The second dataset is GazeBaseVR (GBVR) \cite{lohr2023gazebasevr}, a GazeBase-inspired dataset collected with an eye-tracking-enabled virtual reality (VR) headset.
It includes 5020 binocular recordings from a diverse population of 407 college-aged subjects. 
The data was collected over 26 months in three rounds (Round 1 to Round 3). 
All the eye movements were recorded at a 250~Hz sampling rate.
Each recording captures both horizontal and vertical movements of both eyes in degrees of visual angle. 
Each participant completed a series of five eye movement tasks: vergence (VRG), horizontal smooth pursuit (PUR), reading (TEX), video-viewing (VD), and a random saccade task (RAN).
More details about the dataset and how data were collected are available in \cite{lohr2023gazebasevr}.

\subsection{Model Architecture and Data Handling}
\subsubsection{Data Preprocessing}
All the recordings from each dataset underwent a series of pre-processing steps before being input to the neural network architecture.
EyeLink 1000 is unable to estimate gaze during blinks. 
In these instances, the device returns a Not a Number (NaN) for the affected samples. Additionally, the range of possible horizontal and vertical gaze coordinates is limited to -23.3° to +23.3° and -18.5° to 11.7°, respectively. 
Any gaze samples that fell outside these bounds were also set to NaN.
Two velocity channels (horizontal and vertical) were derived from the raw gaze data using a Savitzky-Golay differentiation filter \cite{savitzkyGolayM} with a window size of 7 and order of 2 \cite{friedman2017method}. 
Subsequently, the recordings were segmented into non-overlapping 5-second sequences (5000-sample) using a rolling window method. 
For each task, the first twelve of these 5-second sequences were then combined into a single 60-second data stream for further analysis.
To mitigate the impact of noise on the data, velocity values were clamped between ±1000°/s. 
Finally, all velocity channels across all sequences and subjects were standardized using z-score normalization.  
In other words, all velocity data from every sample from every sequence and every subject was combined into a single distribution.  
The mean of this distribution was subtracted from every sample, and every sample was divided by the standard deviation of this distribution.  
Any remaining NaN values were replaced with 0, as recommended by Lohr et al. \cite{lohrTBIOM}. 
Further details regarding data pre-processing are provided in Lohr et al. \cite{lohr2022eye}.

\subsubsection{Network Architecture}
In this research, we used the Eye Know You Too (EKYT) network architecture for eye-movement-based biometric authentication\cite{lohr2022eye}. 
This DenseNet-based \cite{densenet} architecture achieves SOTA biometric authentication using high-quality eye-tracking input data (collected at 1000~Hz).
The EKYT architecture incorporates eight convolutional layers. 
In this design, the feature maps generated by each convolutional layer are concatenated with those from all preceding layers before advancing to the subsequent convolutional layer. 
This process results in a concatenated set of feature maps, which are subsequently flattened.
These flattened maps undergo processing through a global average pooling layer and are subsequently input into a fully connected layer, resulting in a 128-dimensional embedding of the input sequence.
The 128-dimensional embedding generated by this architecture serves as the fundamental component for our analysis in this research.
For a more comprehensive understanding of the network architecture, readers are directed to Lohr et al. (2022) \cite{lohr2022eye}.

\subsubsection{Dataset Split \& Training}
For the GB dataset, there were 322 participants in Round 1 and 59 participants in Round 6. 
All 59 participants from Round 6 are a subset of all subjects in Round 1. 
All the participants (59) common through Round 1 to 6 were treated as a held-out dataset and not used for the training and validation step.
The model underwent training using all data (except for heldout data) from Rounds 1-5, except the BLG (gaming) task.

We have used data from all three rounds of the GBVR dataset. 
Round 1 contained data from 407 subjects.
To enhance the integrity of our validation process, we segregated the data of 60 subjects from Round~1 and treated it as a held-out dataset. 
This subset was not used in the training or validation phases.

We divided the participants from training data into four non-overlapping folds for cross-validation. 
The goal was to distribute the number of participants and recordings as evenly as possible across folds. 
The algorithm used for assigning folds is discussed in \cite{lohrTBIOM}.

Four distinct models were trained, with each model using a different fold for validation and the remaining three folds for training.
For learning rate scheduling, we used the Adam\cite{adam} optimizer, and PyTorch's OneCycleLR with cosine annealing \cite{Lr} in the training process.
We used the weighted sum of categorical cross-entropy loss (CE) and multi-similarity loss (MS) loss~\cite{Wang2019} in the training procedure.
We adhered to the default values for the hyperparameters of the MS loss and other optimizer hyperparameters as recommended in \cite{lohr2022eye}.

Our input samples had both horizontal and vertical velocity channels. 
In both the GB and GBVR datasets, the duration for each input sample was set to five seconds. 
Given that GB was collected at a sampling rate of 1000~Hz, each input sample in this dataset includes a window encompassing 5000 time steps. 
Conversely, for the GBVR dataset, which has a sampling rate of 250~Hz, each input sample comprises 1250 time steps.
The model was trained over 100 epochs. 
During the initial 30 epochs, the learning rate was gradually increased to $10^{-2}$ from the initial rate of $10^{-4}$. 
In the subsequent 70 epochs, the learning rate was gradually reduced to a minimum of $10^{-7}$. Each batch contained 64 samples (classes per batch = 8 $\times$ samples per class per batch = 8).

\subsubsection{Embedding Generation}
The method focused on creating centroid embeddings by averaging multiple subsequence embeddings from the first `n' windows of a recording. 
Although the model was not trained directly on these centroid embeddings, it was designed to foster a well-clustered embedding space, ensuring that embeddings from the same class are closely grouped and distinct from others.
The primary process involved embedding the first 5-second sequence of an eye-tracking signal, with the possibility of aggregating embeddings across multiple sequences. 
The training phase did not exclude subsequences with high NaNs, and each subsequence was treated individually.

In our approach, we formed the enrollment dataset by using the first 60 seconds of the session-1 TEX task from Round 1 for each subject in the test set. 
For the authentication dataset, we used the first 60 seconds of the session-2 TEX task from Round 1 for each subject in the test set.
It is to be noted that we did not use 60 seconds at once, we split 60 seconds into 5-second subsequences, getting embeddings for each subsequence, and then computed the centroid of those embeddings.
For each sequence in the enrollment and authentication sets, 128-dimensional embeddings were computed with each of four different models trained using 4-fold cross-validation. 
For simplicity, we are using 128-dimensional embeddings generated from a single-fold model in our study. 
This model was then used to compute pairwise cosine similarities between the embeddings in different sets.

\subsection{Evaluation Metrics}
This study focuses on only three key metrics that assess: (1) the temporal persistence of embeddings, (2) the intercorrelation of embeddings, and (3) biometric performance.

Recall that data from each subject was collected twice in two sessions approximately 20 min apart.  
To assess the temporal persistence of embeddings across sessions, we initially considered employing the Intraclass Correlation Coefficient (ICC) as suggested by \cite{friedman2017method}.
However, most, but not all of the embeddings were normally distributed. For this reason, we opted for the non-parametric Kendall's Coefficient of Concordance (KCC) \cite{field2005kendall} instead of the ICC. 
Intercorrelations between embeddings were assessed using  Spearman correlation coefficient (Spearman R).

Biometric authentication performance was assessed using the equal error rate (EER).  
The EER is the location on a receiver operating characteristic (ROC) curve where the False Rejection Rate and False Acceptance Rate are equal. 
The lower the EER value, the better the performance of the biometric system is. 

The goal of our analysis is to assess the relationships between these three metrics.

\subsection{Hardware \& Software}
All models were trained on a workstation equipped with an NVIDIA RTX A6000 GPU, an AMD Ryzen Threadripper PRO 5975WX with 32 cores, and 48 gigabytes of RAM. The system ran an Anaconda environment with Python 3.7.11, PyTorch 1.10.0, Torchvision 0.11.0, Torchaudio 0.10.0, Cudatoolkit 11.3.1, and Pytorch Metric Learning (PML)~\cite{Musgrave2020a} version 0.9.99.

\section{Experiment Design}

We have designed our experiments based on the research question mentioned in the introduction.

\begin{figure*}[htbp]
\centering
\includegraphics[width=0.85\textwidth]{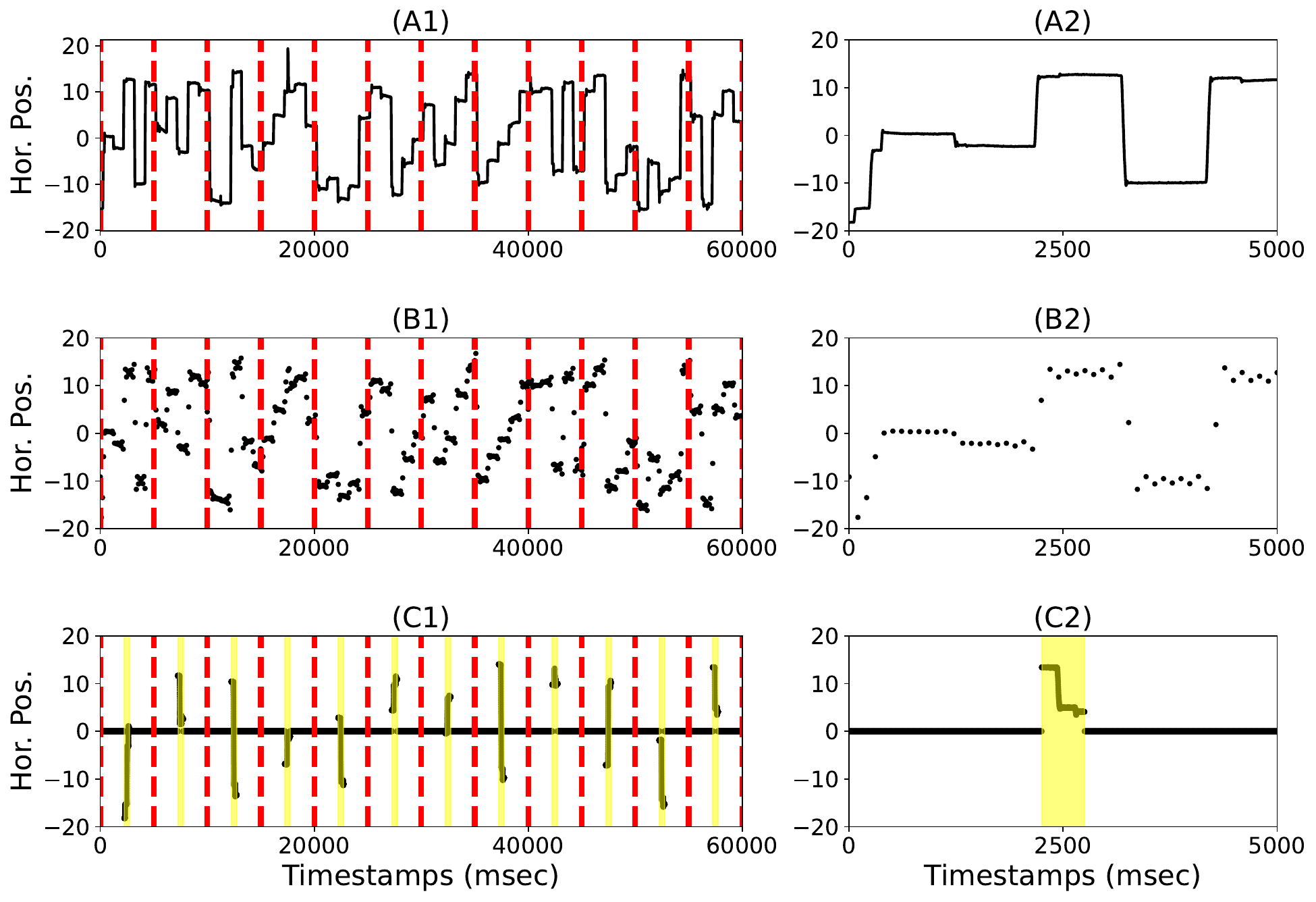}
\caption{Visual Representation of the Experimental Design. 
(A1) The interval between the red-dotted lines is defined as a sequence, containing 5000 samples for GB (for GBVR it is 1250 samples, not shown in the figure). 
(A2) Displays the first sequence from plot (A1). 
(B1) The signal from the plot (A1) has been downsampled to 25 Hz for demonstration. 
(B2) Shows the first sequence from plot (B1). 
(C1) Analyze only the first 10\% of the signal, but place it in the center of the sequence with zero-padding on both sides. 
(C2) Presents the last sequence from the plot (C1) as an example. The right column provides a clearer visualization of the specific sequences from each row.}
\label{fig:ex_sampling}
\end{figure*}

\subsection{\textbf{RQ1: Decimation} Do reductions in the sampling rate affect biometric
performance, and are these changes related to the reliability of the learned embeddings?}

As noted above, GB was initially collected at a sampling frequency of 1000~Hz.
For this analysis, we compared data collected at 1000~Hz to data decimated to frequencies of 500~Hz, 333~Hz, 250~Hz, 100~Hz, 50~Hz, 25~Hz, and 10~Hz. 
GBVR was initially collected at a frequency of 250~Hz, the data was subsequently decimated to frequencies of 125~Hz, 50~Hz, 25~Hz, and 10~Hz.
The decimation process was carried out using the \texttt{scipy.signal.decimate} function, which downsamples the signal after implementing an anti-aliasing filter.
The model in use was then trained on these decimated datasets to produce 128-dimensional embeddings at each decimation level.
Following this, Kendall's Coefficient of Concordance (KCC), and Equal Error Rate (EER) were calculated based on the model and the generated embeddings. Readers are referred to Fig \ref{fig:ex_sampling} (B1-B2).

\subsection{\textbf{RQ2: Percentage} Does reduced data length at a fixed sampling rate affect the reliability of the learned embeddings?}

In the baseline case, each 5-second sequence consisted of 5,000 (GB) or 1,250 (GBVR) samples.  
In this study, we reduced the number of samples used for each sequence by specified percentage levels. 
Each sequence was initially regarded as 100\% of the data. 
We progressively reduced this amount to 50\%, 33\%, 25\%, 10\%, 5\%, 2.5\%, and 1\% for the GB dataset, and to 50\%, 20\%, 10\%, and 4\% for the GBVR dataset. 
Each reduction was applied within a sequence.
For example, for a 50\% reduction, we retained only the first 2.5 seconds of data and zero-padded the rest. 
However, the eye-tracking data was always centered in each sequence because convolutional layers tend to have an effective receptive field that is Gaussian in shape \cite{luo2016understanding}.  
That is, zero-padding was applied to both sides of the reduced data to make the sequence of 5 seconds again. 
For a clearer understanding, readers are referred to Figure \ref{fig:ex_sampling} (C1-C2). 
The model was subsequently trained on these adjusted datasets to produce embeddings. 
We then evaluated the EER, KCC, and intercorrelation based on the model and the generated embeddings.

\subsection{\textbf{RQ3: \# Sequences} Does the number of sequences affect the reliability of the learned embeddings?}

In the baseline model, for GB we averaged embeddings over 12 consecutive 5-second sequences and for GBVR we averaged over 9 consecutive 5-second sequences.  
Our model generates 128-dimensional embeddings for each sequence. For the GB dataset, we evaluated results based on 1 to 12 sequences, and for the GBVR dataset, we evaluated results based on 1 to 9 sequences because of the limited amount of data.
We then analyzed key metrics such as KCC, EER, and the intercorrelation among embeddings derived from differing numbers of sequences.
Refer to Fig \ref{fig:ex_sampling} (A1-A2) for an infographic representation of the experiment.

\subsection{\textbf{RQ4: Signal Quality} Does the quality of the eye-movement signal affect
the reliability of the learned embeddings?}

In this experiment, we inject Gaussian noise into the raw data to downgrade its spatial precision following \cite{prasse, david2021privacy, lohr2022eye}.

We have calculated the precision of the individual recordings.
The raw recording was segmented into 80 ms segments, as referenced in \cite{lohr2019evaluating}. 
Segments containing any invalid samples were excluded. 
We computed the Root Mean Square (RMS) for each valid segment. 
The spatial precision for each recording was determined by calculating the median RMS, considering only the lowest fifth percentile of all RMS values for that recording. 
We then calculated the spatial precision for each subject by taking the median of the spatial precision values from each of their recordings.

Table \ref{tab:hor_rms} shows the spatial precision of GB and GBVR after injecting various amounts of noise.

\begin{table}[htbp]
\centering
\caption{Spatial precision of GB and GBVR at different levels of noise addition.}
\begin{tabular}{ccc}
\toprule
\textbf{Added Noise (SD)} & \textbf{Sp. Precision (GB)} & \textbf{Sp. Precision (GBVR)} \\ 
\midrule
0 & 0.0044 & 0.041 \\ 
0.05 & 0.059 & 0.070 \\ 
0.25 & 0.289 & 0.240 \\ 
0.5 & 0.577 & 0.460 \\ 
0.75 & 0.865 & 0.683 \\ 
1.0 & 1.151 & 0.905 \\ 
1.25 & 1.438 & 1.129 \\ 
1.5 & 1.726 & 1.351 \\ 
1.75 & 2.013 & 1.576 \\ 
2 & 2.301 & 1.796 \\ 
\bottomrule

\end{tabular}
\label{tab:hor_rms}
\end{table}

\subsection{\textbf{RQ5: Intercorrelation} Does any of the eye-tracking data manipulation affect the intercorrelation of the learned embeddings?}

We have calculated the absolute value of the intercorrelation (using \texttt{scipy.stats.spearmanr}) for each of the above analyses. We have investigated the effect of the eye-tracking data manipulation on the calculated intercorrelation.

\section{Results}

In this study, we employ four manipulations of eye-movement data (decimation, percentage, number of sequences, and signal quality degradation) and evaluate the effects of these manipulations on biometric performance (EER), temporal persistence (KCC) and the relationship between EER and KCC.  We also evaluate the intercorrelations of embeddings after each manipulation.

For RQ1, RQ2 and RQ3, we found that the relationships between the manipulation and either biometric performance or reliability (KCC) were the best fit after taking the log of x, whereas for RQ4, we found that a linear fit was best. The following equations are used.

\begin{table}[htbp]
\centering
\begin{tabular}{cc}
\toprule
\textbf{Linear Fit}  & \textbf{Logarithmic Fit} \\
\midrule
f(x) = ax + b  & f(x) = $a \cdot \log(x) + b$ \\

\multicolumn{1}{>{\raggedright\arraybackslash}m{3.8cm}}{\textbf{Where:}} 
&
\multicolumn{1}{>{\raggedright\arraybackslash}m{3.8cm}}{\textbf{Where:}}
\\

\multicolumn{1}{>{\raggedright\arraybackslash}m{3.8cm}}{\( x \) is the independent variable,}  & 
\multicolumn{1}{>{\raggedright\arraybackslash}m{3.8cm}}{\( x \) is the independent variable,}
\\

\multicolumn{1}{>{\raggedright\arraybackslash}m{3.8cm}}{\( a \) is the slope of the line,} 
&
\multicolumn{1}{>{\raggedright\arraybackslash}m{3.8cm}}{\( a \) is the slope of the logarithmic curve,}
\\

\multicolumn{1}{>{\raggedright\arraybackslash}m{3.8cm}}{\( b \) is the y-intercept of the line.} & \multicolumn{1}{>{\raggedright\arraybackslash}m{3.8cm}}{\( b \) is the y-intercept of the logarithmic curve.} \\

\bottomrule
\end{tabular}
\end{table}

\subsection{RQ1: Do reductions in the sampling rate affect biometric
performance, and are these changes related to the reliability of the learned embeddings?}

\begin{figure}[htbp]
\centering
\includegraphics[width=0.9\textwidth]{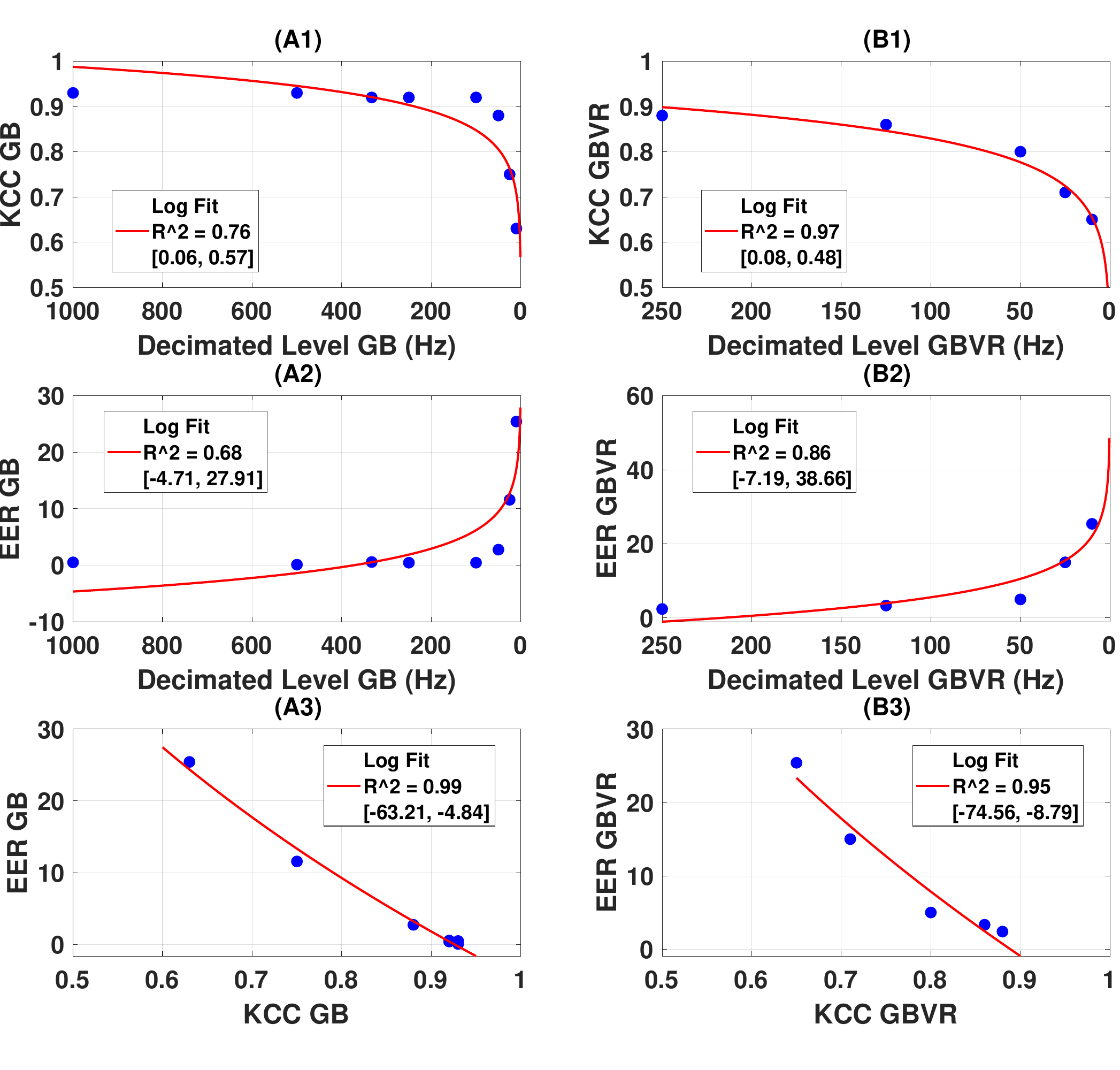}
\caption{Relationship between KCC and EER with the decimated level (Hz) for two datasets: GB and GBVR. 
Subplots (A1) and (B1) show the logarithmic decrease in KCC for the GB and GBVR datasets, respectively, with lower sampling rates. 
Subplots (A2) and (B2) depict the logarithmic increase in EER for the GB and GBVR datasets, with lower sampling rates. 
Subplots (A3) and (B3) illustrate the strong negative logarithmic relationship between EER and KCC for the GB and GBVR datasets.
$R^2$, and coefficient values are added to each plot's legend.
The $R^2$ values across all plots indicate a strong fit, suggesting that a higher sampling rate improves biometric performance, with the GBVR dataset demonstrating a particularly robust fit.}
\label{fig:decimate}
\end{figure}

Figure \ref{fig:decimate} illustrates how sampling rate affects the temporal persistence and the performance of the EMB system, comparing two different datasets. The provided figure illustrates the relationships between two performance metrics (KCC and EER) and the decimated level (in Hz) for two datasets: GB and GBVR. 
The left column of plots shows that KCC decreases with a decreasing sampling rate for the GB dataset, while EER increases logarithmically, indicating that a lower sampling rate negatively impacts performance.

This is evidenced by $R^2$ values of 0.76 and 0.68 for KCC and EER, respectively, and a strong negative correlation between EER and KCC with a  $R^2$ value of 0.99. 
Similarly, the right column reveals that for the GBVR dataset, KCC also decreases and EER increases with a lower sampling rate, but with stronger fits ($R^2$ values of 0.97 and 0.86 for KCC and EER, respectively) and nearly perfect negative correlation between EER and KCC ($R^2$ value of 0.95). 
Overall, the figure indicates that higher sampling rates yield better temporal persistence and lower equal error rates in EMB, with GBVR appearing more sensitive to changes in sampling rate than GB. 
The logarithmic models show strong correlations, emphasizing the importance of maintaining high sampling rates for better biometric performance.

Decimation reduces both data length and sampling rate. So, after decimation, it is impossible to determine if the effects on our measures is due to the reduced sample rate or the reduced amount of data.  To address this confound, we also performed the percentage analysis.  For this, the same number of eye-movement signal points is used as for decimation but the sampling rate does not change. 


\subsection{RQ2: Does reduced data length at a fixed sampling rate affect the reliability of the learned embeddings?}

\begin{figure}[htbp]
\centering
\includegraphics[width=0.9\textwidth]{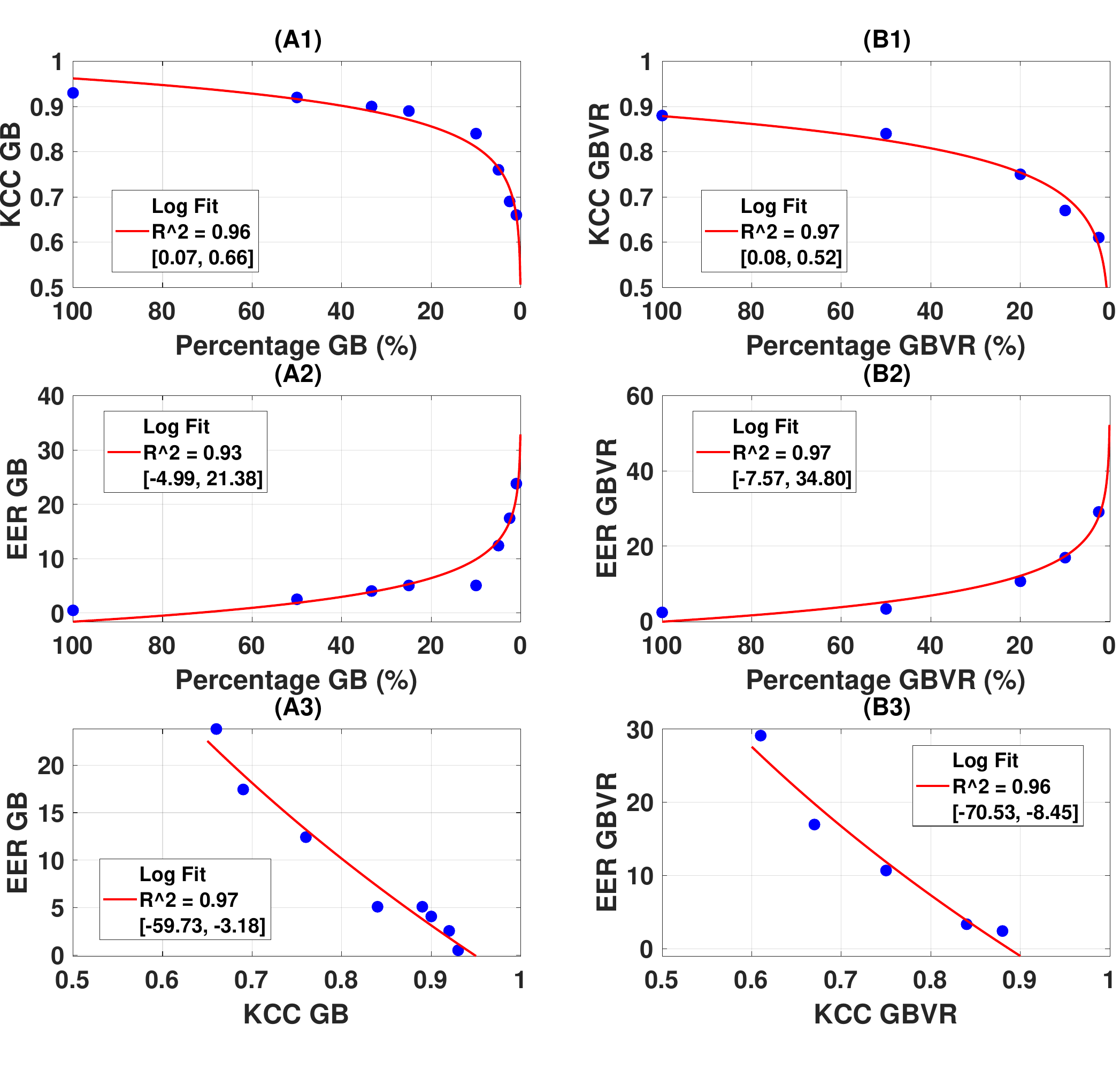}
\caption{
Relationship between KCC and EER with the percentage level (\%) for two datasets: GB and GBVR. 
Subplots (A1) and (B1) show the logarithmic decrease in KCC for the GB and GBVR datasets, respectively, with lower percentage levels. 
Subplots (A2) and (B2) depict the logarithmic increase in EER for the GB and GBVR datasets as the percentage levels decrease.  The high $R^2$ values across these plots indicate a strong fit, suggesting that a higher percentage level improves biometric performance.
Subplots (A3) and (B3) illustrate the strong negative logarithmic correlation between EER and KCC for the GB and GBVR datasets. 
$R^2$, and coefficient values are added to each plot's legend.}
\label{fig:percentage}
\end{figure}

Figure \ref{fig:percentage} illustrates how reducing the percentage of raw eye movement data impacts temporal persistence and performance of the EMB system. 
The figure shows the relationships between two performance metrics— KCC and EER — and the percentage level (\%).
In the left column of plots, it is evident that for the GB dataset, KCC decreases with decreasing percentage levels, with an $R^2$ value of 0.96, while EER increases logarithmically, with an  $R^2$ value of 0.93. 
There is a strong negative correlation between EER and KCC, indicated by an $R^2$ value of 0.97. 
Similarly, the right column of plots reveals that for the GBVR dataset, KCC decreases with decreasing percentage levels, with a  $R^2$ value of 0.97, and EER increases logarithmically, with an $R^2$ value of 0.97. 
The negative correlation between EER and KCC is almost perfect, with an $R^2$ value of 0.96. 
Overall, the figure suggests that a higher percentage of eye movement data leads to better temporal persistence and lower equal error rates in the EMB system.

\begin{figure}[htbp]
\centering
\includegraphics[width=0.85\textwidth]{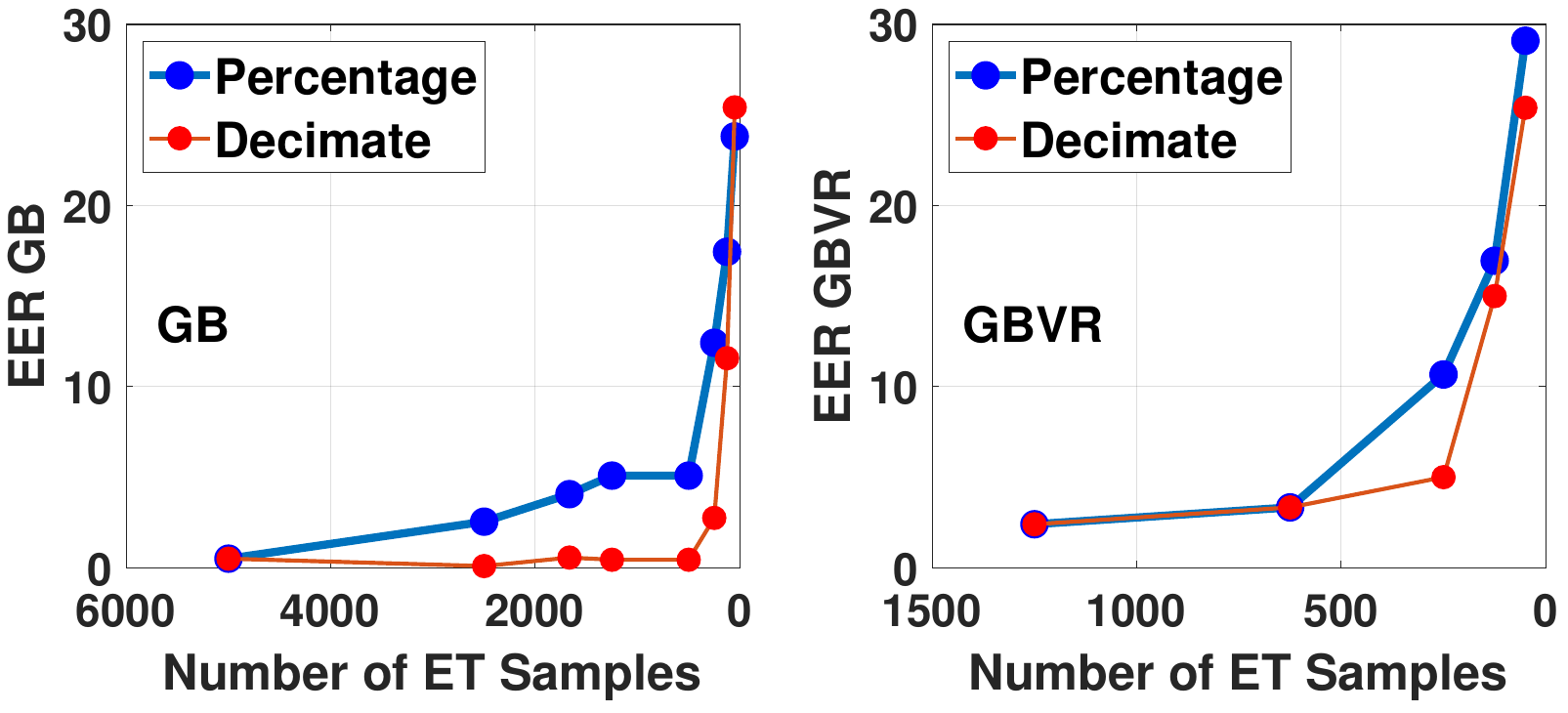}
\caption{Relationship between the number of eye-tracking (ET) samples and the Equal Error Rate (EER). The plots illustrate the EER for two datasets, GB and GBVR, across varying sample sizes (50 to 5000 samples for GB and 50 to 1250 samples for GBVR). The results from RQ1 and RQ2 have been compared in each plot.}
\label{fig:EER_Plots}
\end{figure}

In Fig. \ref{fig:EER_Plots}, we compare the biometric performance between decimation and percentage analysis. In the GB plot, for decimation, we can see the biometric performance degrades significantly when the number of samples is reduced to 250. On the other hand, for percentage analysis, the biometric performance degrades from the very beginning.
A similar trend is observed in the GBVR plot, though the difference between decimation and percentage is more subtle.

\subsection{RQ3: Does the number of sequences affect the reliability of the learned embeddings?}

\begin{figure}[htbp]
\centering
\includegraphics[width=0.9\textwidth]{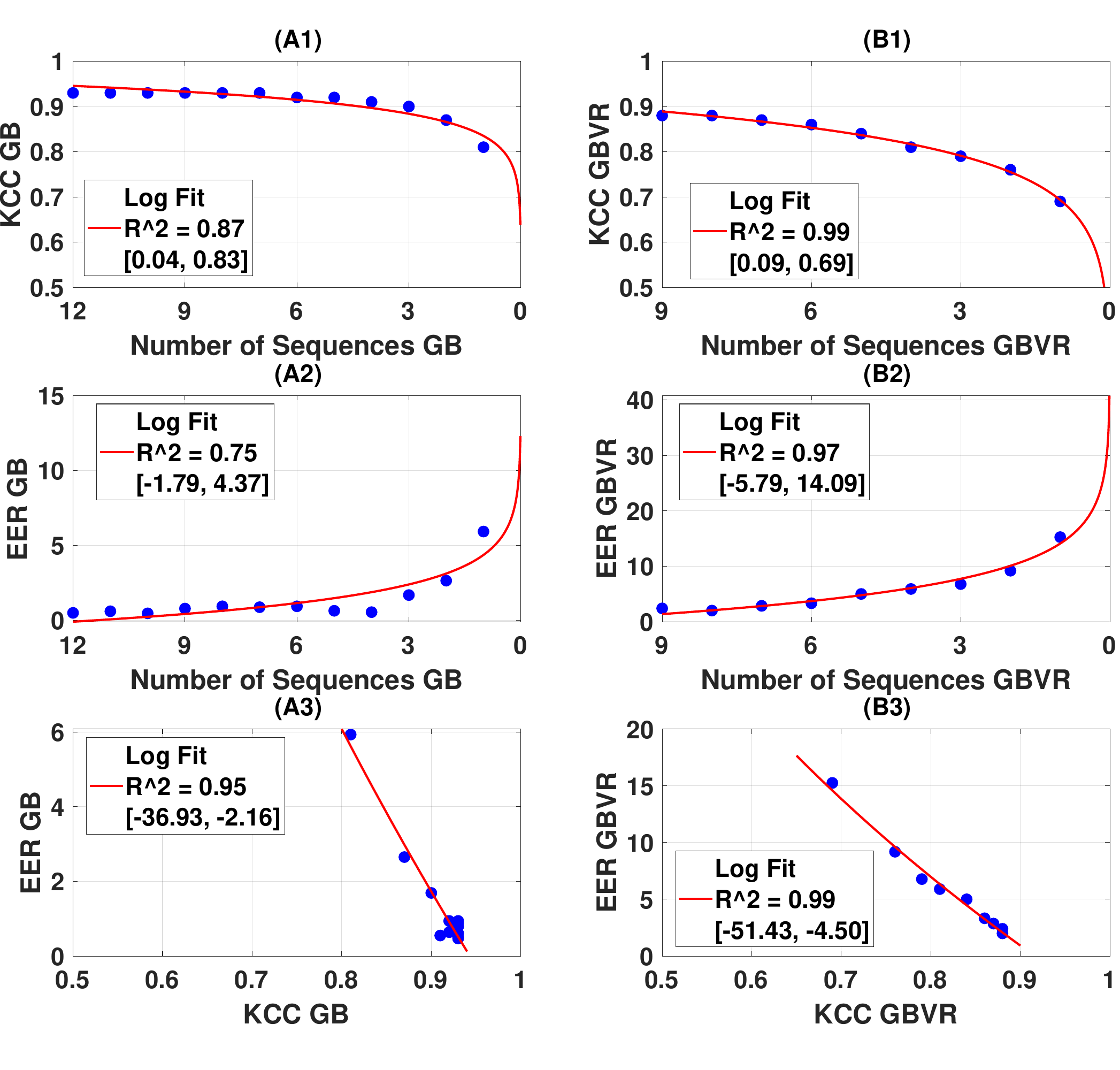}
\caption{Relationship between KCC and EER with the number of sequences for two datasets: GB and GBVR. 
Subplots (A1) and (B1) show the logarithmic decrease in KCC GB and KCC GBVR, respectively, with reduced sequences. 
Subplots (A2) and (B2) depict the logarithmic increase in EER GB and EER GBVR with increasing sequences. 
Subplots (A3) and (B3) illustrate the strong negative logarithmic correlation between EER and KCC for GB and GBVR datasets, presenting higher temporal persistence associated with lower equal error rates. 
$R^2$, and coefficient values are added to each plot's legend.}
\label{fig:datalength}
\end{figure}

Figure \ref{fig:datalength} illustrates the impact of the number of sequences (GB: 5,000 samples, GBVR: 1,250 samples) on the temporal persistence of the learned embeddings and the performance of the EMB system.  It consists of six subplots showcasing the relationships between two performance metrics (KCC and EER) and the number of sequences, both for GB and GBVR datasets.

The left column focuses on the GB dataset, illustrating how KCC and EER change with decreasing number of sequences. 
The KCC shows a logarithmic decrease with a decrease in the number of sequences (A1) with an $R^2$ value of 0.87, indicating a strong fit. 
Conversely, EER increases with less number of sequences (A2), supported by an $R^2$ value of 0.75. 
The relationship between KCC and EER (A3) reveals a strong negative logarithmic correlation, with an $R^2$ value of 0.95, indicating a strong fit.
The right column mirrors these analyses for the GBVR dataset, where KCC also decreases with the number of sequences reduced (B1) and EER increases (B2), with $R^2$ values of 0.99 and 0.97. 
The EER vs. KCC plot (B3) shows a negative relationship with an $R^2$ value of 0.99. 
Overall, these results indicate that biometric performance improves with more sequences, reflected in better reliability (KCC) and lower error rates (EER), with the GBVR dataset demonstrating an even more robust fit than the GB dataset alone.


\subsection{RQ4: Does the quality of the eye-movement signal affect
the reliability of the learned embeddings?}

\begin{figure}[htbp]
\centering
\includegraphics[width=0.9\textwidth]{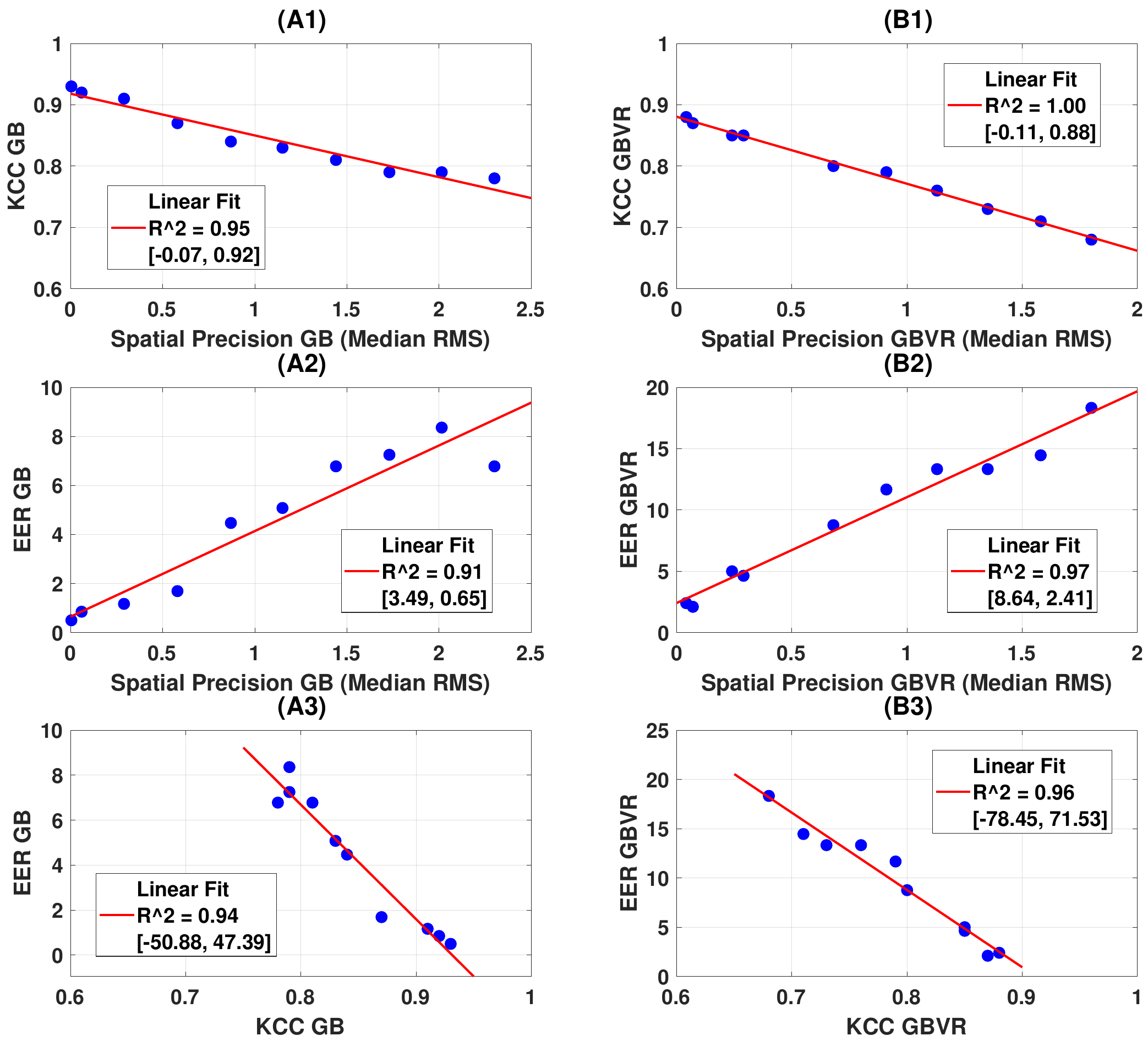}
\caption{Relationship between KCC and EER with the variation in spatial precision for two datasets: GB and GBVR. 
Subplots (A1) and (B1) show the linear decrease in KCC GB and KCC GBVR, respectively, with the degradation of spatial precision. 
Subplots (A2) and (B2) depict the linear increase in EER GB and EER GBVR with the degradation of spatial precision. 
Subplots (A3) and (B3) illustrate the strong negative linear correlation between KCC and EER for GB and GBVR datasets, presenting higher temporal persistence associated with lower equal error rates. 
$R^2$, and coefficient values are added to each plot's legend.}
\label{fig:NoiseLinear}
\end{figure}

Fig. \ref{fig:NoiseLinear} shows how degraded spatial precision affects KCC and EER, across embeddings learned from two datasets.
For the GB dataset, spatial precision values range from 0.00435 (original) to 2.3 with the injection of Gaussian noise.
For the GBVR dataset, spatial precision values range from 0.041(original) to 1.80 with the injection of Gaussian noise.
A significant relationship is observed in both GB and GBVR; KCC drops and EER increases linearly with degraded spatial precision.
A strong negative linear correlation exists between KCC and EER for GB and GBVR datasets, presenting higher temporal persistence associated with lower equal error rates.

\subsection{RQ5: Does any of the eye-tracking data manipulation affect the intercorrelation of the learned embeddings?}
We calculated the absolute value of the intercorrelation of the learned embeddings for each of the analyses above and found that data manipulation minimally affects intercorrelation. When combining all levels of manipulation, across all datasets, the mean absolute correlation value is 0.19 with an SD of 0.14. Detailed results are shown in Table. \ref{tab:intercorr}.

\begin{table}[htbp]
    \centering
    \caption{Absolute Value of Intercorrelation}
    \begin{tabular}{lcc}
        \toprule
         Experiment name & Mean (SD) - GB & Mean (SD) - GBVR   \\
         \midrule
         Decimation & 0.19 (0.14) & 0.19 (0.14)\\
         Percentage & 0.20 (0.15) & 0.20 (0.15)\\
         \# Sequences & 0.19 (0.14) & 0.18 (0.14)\\
         Degraded Signal Quality & 0.19 (0.14) & 0.20 (0.15)\\
         \bottomrule
    \end{tabular}
    \label{tab:intercorr}
\end{table}


\subsection{Relation between Temporal Persistence (KCC) and Biometric Performance(EER) across All Manipulations}

We have seen that there is a strong relationship between KCC and EER in each of the above analyses.  The relationship between these two parameters is very strong (Fig. 7, $R^2=0.92$), supporting our notion that temporal persistence (``reliability'') is important for biometric performance in all cases.   

\begin{figure}[htbp]
\centering
\includegraphics[width=0.85\textwidth]{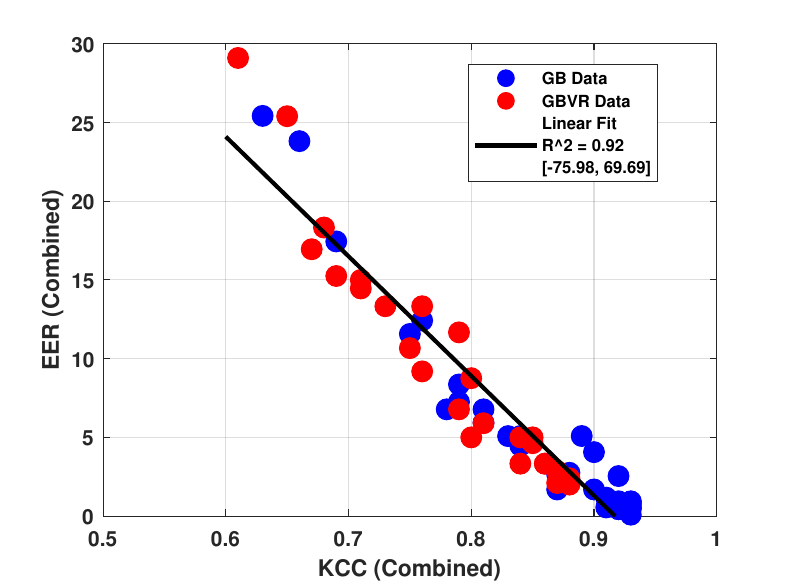}
\caption{This graph compares the relationship between KCC and EER across all manipulations and datasets. A linear model provides a good fit with a model $r^2$ of 0.92.}
\label{fig:kccvsEER}
\end{figure}

\section{Discussion}

The main findings of this report are listed in Table \ref{tab:comparison}.  This table shows the effect of various signal manipulations on embeddings in terms of reliability (KCC), intercorrelation, and biometric performance (EER).  

\begin{table}[htbp]
    \centering
    \caption{Impact on Embeddings: Performance Metrics Variation}
    \begin{threeparttable}
    \begin{tabular}{ccccc}
        \toprule
        \textbf{RQ$\dagger$} & \textbf{Description} & \textbf{KCC$\ddagger$} & \textbf{Spearman R*} & \textbf{EER$\ddagger$} \\
        \midrule
        1 & Decimate & \textcolor{red}{$\times$} & \textemdash & \textcolor{red}{$\times$}  \\ 
        2 & Percentage & \textcolor{red}{$\times$} & \textemdash & \textcolor{red}{$\times$}  \\
        3 & \# Sequences & \textcolor{red}{$\times$} & \textemdash & \textcolor{red}{$\times$} \\ 
        4 & Sig. Quality & \textcolor{red}{$\times$} & \textemdash & \textcolor{red}{$\times$} \\ 
        \midrule
    \end{tabular}
    \begin{tablenotes}
        \small
        \item{ $\dagger$ Research Question}
        \item{$\ddagger$ \textcolor{red}{$\times$} indicates a significant effect due to data manipulation}
        \item{$*$ \textbf{``\textemdash''} Denotes negligible impact.}
    \end{tablenotes}
    \end{threeparttable}
    \label{tab:comparison}
\end{table}


\subsection{Summary of Findings}
\begin{enumerate}
        
\item \textbf{RQ1: Decimation } 
The impact of different sampling rates on embeddings was profound. 
For GB, the experiment showed that while decimation down to 100 Hz did not have a major impact on either stability (KCC) or biometric performance (EER), decimation below this frequency led to marked drops in reliability and decreases in biometric performance.  
For GBVR, the point of transition was closer to 50 Hz.  
This suggests that while some degree of decimation is tolerable, excessively low sampling rates compromise the efficacy of the biometric system to a significant degree.
   
\textbf{RQ2: Percentage } We have found that reduced data length from the recording during the training process significantly influenced the embeddings. 
Lowering the given data percentages resulted in less reliable embeddings, as evidenced by a downward trend of KCC from 100\% to 1 \% of data. 
A significant drop in biometric performance in terms of Equal Error Rate (EER) is also seen in the results.

When we compare the effect of decimation to the effect of a reduced percentage of samples, we note that biometric performance is better with decimation than with the reduction of data samples alone.  This is probably because the decimated signal still samples the entire signal whereas the percentage manipulation only samples a small part of the signal.

\item \textbf{RQ3: \#Sequences}  We found that varying sequence sizes significantly influenced the embeddings. 
Longer sequences generally resulted in more reliable and consistent embeddings, as evidenced by higher KCC values for 12 consecutive sequences (60 seconds) compared to 1 sequence (5 seconds) of data. 
However, there was a diminishing return on increasing data length beyond a certain point. 
The most significant impact was observed when comparing very short data sequences to moderately long ones. 
The improvement in embedding reliability was marked, as evidenced by a noticeable shift in key metrics.
A significant drop in biometric performance in terms of EER is also seen in the results. 

\item \textbf{RQ4: Signal Quality} The study also delved into how eye-tracking signal quality (spatial precision) affects embeddings. We noted its influence on the embeddings' KCC and EER. 
Results indicated that there is a significant effect on temporal persistence and biometric performance with downgrading spatial precision eye-tracking signal by injecting Gaussian noise. 

\item \textbf{RQ5: Intercorrelation} For all analyses, the effect of any of our manipulations had a minimal effect on the absolute value of the intercorrelations of the embeddings.  In all cases, the absolute value of the intercorrelations was small (mean = 0.19, SD = 0.14).
    
\end{enumerate}

It was interesting to note that when the relationship between KCC and EER was evaluated across all manipulations and both datasets, the relationship was very strong.  Thus, it appears that there may be a universal strong relationship between these two entities.

\subsection{Methodological Strength}
A significant methodological strength of our study lies in the selection of the Eye Know You Too (EKYT) network architecture, which is recognized for its state-of-the-art performance in biometric authentication 
The EKYT architecture is based on a DenseNet framework, known for its efficient handling of complex data structures.

Additionally, we designed and conducted a series of experiments to manipulate the quality of eye movement signals. These experiments included altering the sampling rate, reducing the sample size, varying the number of sequence sizes, and degrading the spatial precision of the signal. These manipulations allowed us to thoroughly investigate the robustness of the relation between the temporal persistence and biometric performance of the DL-based EMB.

\subsection{Limitations \& Future Direction}
All of our biometric measurements emerge from an ROC analysis.  It is well established that ROC-curve analyses are relatively inaccurate when based on small sample sizes \cite{smallsample}.  For GB, all biometric performance was assessed with N=59 subjects, and for GBVR, all biometric performance was assessed with N = 60 subjects.  These are small samples.  However, in the context of eye-tracking studies, our sample sizes are relatively large. Even with such small sample sizes, we did find important relationships among the variables of interest.  However, a replication of this work with larger sample sizes would make an important contribution.  


Obviously, good biometric performance results from embeddings with high temporal persistence.
A key question arises for future research: Can the temporal persistence of embeddings be integrated into the overall biometric analysis? In other words, is there a method to enhance the temporal persistence of embeddings learned by a DL-based biometric pipeline?

\section{Conclusion}

We have previously shown the importance of temporal persistence on biometric performance in traditional, non-DL based biometric systems \cite{friedman2017method}.  
The findings in our present report extend this finding to embeddings learned by a DL-based biometric pipeline.  
Our study has shown a strong relation between the temporal persistence of learned embeddings as assessed by the KCC to the biometric performance (EER) of a DL-based biometric pipeline.
We have also documented the effects of various data manipulations on biometric performance and temporal persistence.  
Data manipulation in any manner affects the learned embeddings from the temporal persistence and biometric efficacy perspective. 
Intercorrelations do not vary much throughout the conducted research.





\bibliographystyle{bib/IEEEtran}
\bibliography{bib/IEEEabrv,01main}

\begin{thebibliography}{10}
\providecommand{\url}[1]{#1}
\csname url@samestyle\endcsname
\providecommand{\newblock}{\relax}
\providecommand{\bibinfo}[2]{#2}
\providecommand{\BIBentrySTDinterwordspacing}{\spaceskip=0pt\relax}
\providecommand{\BIBentryALTinterwordstretchfactor}{4}
\providecommand{\BIBentryALTinterwordspacing}{\spaceskip=\fontdimen2\font plus
\BIBentryALTinterwordstretchfactor\fontdimen3\font minus \fontdimen4\font\relax}
\providecommand{\BIBforeignlanguage}[2]{{%
\expandafter\ifx\csname l@#1\endcsname\relax
\typeout{** WARNING: IEEEtran.bst: No hyphenation pattern has been}%
\typeout{** loaded for the language `#1'. Using the pattern for}%
\typeout{** the default language instead.}%
\else
\language=\csname l@#1\endcsname
\fi
#2}}
\providecommand{\BIBdecl}{\relax}
\BIBdecl

\bibitem{alrawili2023comprehensive}
R.~Alrawili, A.~A.~S. AlQahtani, and M.~K. Khan, ``Comprehensive survey: Biometric user authentication application, evaluation, and discussion,'' \emph{arXiv preprint arXiv:2311.13416}, 2023.

\bibitem{Nelson}
\BIBentryALTinterwordspacing
J.~{Nelson, CPP}, ``Chapter 12 - biometrics characteristics,'' in \emph{Effective Physical Security (Fourth Edition)}, 4th~ed., L.~J. Fennelly, Ed.\hskip 1em plus 0.5em minus 0.4em\relax Butterworth-Heinemann, 2013, pp. 255--256. [Online]. Available: \url{https://www.sciencedirect.com/science/article/pii/B9780124158924000122}
\BIBentrySTDinterwordspacing

\bibitem{van2013characteristics}
H.~van~de Haar, D.~van Greunen, and D.~Pottas, ``The characteristics of a biometric,'' in \emph{2013 Information Security for South Africa}.\hskip 1em plus 0.5em minus 0.4em\relax IEEE, 2013, pp. 1--8.

\bibitem{harvey2017permanence}
J.~Harvey, J.~Campbell, S.~Elliott, M.~Brockly, and A.~Adler, ``Biometric permanence: Definition and robust calculation,'' in \emph{2017 Annual IEEE International Systems Conference (SysCon)}, 2017, pp. 1--7.

\bibitem{das2022advancing}
P.~Das, ``Advancing the state-of-the-art in iris biometrics: Permanence, individuality and security,'' Ph.D. dissertation, Clarkson University, 2022.

\bibitem{bargary2017individual}
\BIBentryALTinterwordspacing
G.~Bargary, J.~M. Bosten, P.~T. Goodbourn, A.~J. Lawrance-Owen, R.~E. Hogg, and J.~D. Mollon, ``\BIBforeignlanguage{en}{Individual differences in human eye movements: {An} oculomotor signature?}'' \emph{\BIBforeignlanguage{en}{Vision Research}}, vol. 141, pp. 157--169, Dec. 2017. [Online]. Available: \url{http://www.sciencedirect.com/science/article/pii/S0042698917300391}
\BIBentrySTDinterwordspacing

\bibitem{friedman2017method}
\BIBentryALTinterwordspacing
L.~Friedman, M.~S. Nixon, and O.~V. Komogortsev, ``Method to assess the temporal persistence of potential biometric features: Application to oculomotor, gait, face and brain structure databases,'' \emph{PLOS ONE}, vol.~12, no.~6, pp. 1--42, 06 2017. [Online]. Available: \url{https://doi.org/10.1371/journal.pone.0178501}
\BIBentrySTDinterwordspacing

\bibitem{Temporal}
L.~Friedman, H.~S. Stern, L.~R. Price, and O.~V. Komogortsev, ``Why temporal persistence of biometric features, as assessed by the intraclass correlation coefficient, is so valuable for classification performance,'' \emph{Sensors}, vol.~20, no.~16, p. 4555, 2020.

\bibitem{kasprowski2004}
\BIBentryALTinterwordspacing
P.~Kasprowski and J.~Ober, ``{Eye movements in biometrics},'' \emph{Lecture Notes in Computer Science (including subseries Lecture Notes in Artificial Intelligence and Lecture Notes in Bioinformatics)}, vol. 3087, pp. 248--258, 2004. [Online]. Available: \url{https://doi.org/10.1007/978-3-540-25976-3_23}
\BIBentrySTDinterwordspacing

\bibitem{katsini2020role}
\BIBentryALTinterwordspacing
C.~Katsini, Y.~Abdrabou, G.~E. Raptis, M.~Khamis, and F.~Alt, ``The {Role} of {Eye} {Gaze} in {Security} and {Privacy} {Applications}: {Survey} and {Future} {HCI} {Research} {Directions},'' in \emph{Proceedings of the 2020 {CHI} {Conference} on {Human} {Factors} in {Computing} {Systems}}, ser. {CHI} '20.\hskip 1em plus 0.5em minus 0.4em\relax Honolulu, HI, USA: Association for Computing Machinery, Apr. 2020, pp. 1--21. [Online]. Available: \url{https://doi.org/10.1145/3313831.3376840}
\BIBentrySTDinterwordspacing

\bibitem{mh2023determining}
\BIBentryALTinterwordspacing
M.~H. Raju, L.~Friedman, T.~Bouman, and O.~Komogortsev, ``Determining which sine wave frequencies correspond to signal and which correspond to noise in eye-tracking time-series,'' \emph{Journal of Eye Movement Research}, vol.~14, no.~3, Dec. 2023. [Online]. Available: \url{https://bop.unibe.ch/JEMR/article/view/9887}
\BIBentrySTDinterwordspacing

\bibitem{mh2023filtering}
\BIBentryALTinterwordspacing
------, ``Filtering eye-tracking data from an eyelink 1000: Comparing heuristic, savitzky-golay, iir and fir digital filters,'' \emph{Journal of Eye Movement Research}, vol.~14, no.~3, Oct. 2023. [Online]. Available: \url{https://bop.unibe.ch/JEMR/article/view/9888}
\BIBentrySTDinterwordspacing

\bibitem{schroder2020robustness}
C.~Schr{\"o}der, S.~M.~K. Al~Zaidawi, M.~H. Prinzler, S.~Maneth, and G.~Zachmann, ``Robustness of eye movement biometrics against varying stimuli and varying trajectory length,'' in \emph{Proceedings of the 2020 CHI Conference on Human Factors in Computing Systems}, 2020, pp. 1--7.

\bibitem{rigas2017current}
I.~Rigas and O.~V. Komogortsev, ``Current research in eye movement biometrics: An analysis based on bioeye 2015 competition,'' \emph{Image and Vision Computing}, vol.~58, pp. 129--141, 2017.

\bibitem{deepeyedentification}
L.~A. J{\"a}ger, S.~Makowski, P.~Prasse, S.~Liehr, M.~Seidler, and T.~Scheffer, ``Deep eyedentification: Biometric identification using micro-movements of the eye,'' in \emph{Machine Learning and Knowledge Discovery in Databases: European Conference, ECML PKDD 2019, W{\"u}rzburg, Germany, September 16--20, 2019, Proceedings, Part II}.\hskip 1em plus 0.5em minus 0.4em\relax Springer, 2020, pp. 299--314.

\bibitem{deepeyedentificationlive}
S.~Makowski, P.~Prasse, D.~R. Reich, D.~Krakowczyk, L.~A. Jäger, and T.~Scheffer, ``Deepeyedentificationlive: Oculomotoric biometric identification and presentation-attack detection using deep neural networks,'' \emph{IEEE Transactions on Biometrics, Behavior, and Identity Science}, vol.~3, no.~4, pp. 506--518, 2021.

\bibitem{lohr2020metric}
\BIBentryALTinterwordspacing
D.~Lohr, H.~Griffith, S.~Aziz, and O.~Komogortsev, ``A metric learning approach to eye movement biometrics,'' in \emph{2020 IEEE International Joint Conference on Biometrics (IJCB)}.\hskip 1em plus 0.5em minus 0.4em\relax IEEE, 2020, pp. 1--7. [Online]. Available: \url{http://dx.doi.org/10.1109/IJCB48548.2020.9304859}
\BIBentrySTDinterwordspacing

\bibitem{Lohr2020}
\BIBentryALTinterwordspacing
D.~J. Lohr, S.~Aziz, and O.~Komogortsev, ``Eye movement biometrics using a new dataset collected in virtual reality,'' in \emph{ACM Symposium on Eye Tracking Research and Applications}, ser. ETRA ’20 Adjunct.\hskip 1em plus 0.5em minus 0.4em\relax New York, NY, USA: Association for Computing Machinery, 2020. [Online]. Available: \url{https://doi.org/10.1145/3379157.3391420}
\BIBentrySTDinterwordspacing

\bibitem{lohrTBIOM}
D.~Lohr, H.~Griffith, and O.~V. Komogortsev, ``Eye know you: Metric learning for end-to-end biometric authentication using eye movements from a longitudinal dataset,'' \emph{IEEE Transactions on Biometrics, Behavior, and Identity Science}, 2022.

\bibitem{lohr2022eye}
D.~Lohr and O.~V. Komogortsev, ``Eye know you too: Toward viable end-to-end eye movement biometrics for user authentication,'' \emph{IEEE Transactions on Information Forensics and Security}, vol.~17, pp. 3151--3164, 2022.

\bibitem{raju2024signal}
M.~H. Raju, L.~Friedman, D.~J. Lohr, and O.~Komogortsev, ``Signal vs noise in eye-tracking data: Biometric implications and identity information across frequencies,'' in \emph{Proceedings of the 2024 Symposium on Eye Tracking Research and Applications}, 2024, pp. 1--7.

\bibitem{nilsson2016screening}
M.~Nilsson~Benfatto, G.~{\"O}qvist~Seimyr, J.~Ygge, T.~Pansell, A.~Rydberg, and C.~Jacobson, ``Screening for dyslexia using eye tracking during reading,'' \emph{PloS one}, vol.~11, no.~12, p. e0165508, 2016.

\bibitem{billeci2017integrated}
L.~Billeci, A.~Narzisi, A.~Tonacci, B.~Sbriscia-Fioretti, L.~Serasini, F.~Fulceri, F.~Apicella, F.~Sicca, S.~Calderoni, and F.~Muratori, ``An integrated eeg and eye-tracking approach for the study of responding and initiating joint attention in autism spectrum disorders,'' \emph{Scientific Reports}, vol.~7, no.~1, p. 13560, 2017.

\bibitem{sargezeh2019gender}
B.~A. Sargezeh, N.~Tavakoli, and M.~R. Daliri, ``Gender-based eye movement differences in passive indoor picture viewing: An eye-tracking study,'' \emph{Physiology \& behavior}, vol. 206, pp. 43--50, 2019.

\bibitem{al2020gender}
S.~M.~K. Al~Zaidawi, M.~H. Prinzler, C.~Schr{\"o}der, G.~Zachmann, and S.~Maneth, ``Gender classification of prepubescent children via eye movements with reading stimuli,'' in \emph{Companion Publication of the 2020 International Conference on Multimodal Interaction}, 2020, pp. 1--6.

\bibitem{eberz2015preventing}
\BIBentryALTinterwordspacing
S.~Eberz, K.~Rasmussen, V.~Lenders, and I.~Martinovic, ``Preventing lunchtime attacks: Fighting insider threats with eye movement biometrics,'' in \emph{Network and Distributed System Security (NDSS) Symposium}.\hskip 1em plus 0.5em minus 0.4em\relax Internet Society, 2015. [Online]. Available: \url{http://dx.doi.org/10.14722/ndss.2015.23203}
\BIBentrySTDinterwordspacing

\bibitem{Komogortsev2015}
\BIBentryALTinterwordspacing
O.~V. {Komogortsev}, A.~{Karpov}, and C.~D. {Holland}, ``Attack of mechanical replicas: {Liveness} detection with eye movements,'' \emph{IEEE Transactions on Information Forensics and Security}, vol.~10, no.~4, pp. 716--725, 2015. [Online]. Available: \url{https://doi.org/10.1109/TIFS.2015.2405345}
\BIBentrySTDinterwordspacing

\bibitem{rigas2015}
\BIBentryALTinterwordspacing
I.~Rigas and O.~V. Komogortsev, ``{Eye Movement-Driven Defense against Iris Print-Attacks},'' \emph{Pattern Recogn. Lett.}, vol.~68, no.~P2, p. 316–326, dec 2015. [Online]. Available: \url{http://dx.doi.org/10.1016/j.patrec.2015.06.011}
\BIBentrySTDinterwordspacing

\bibitem{raju2022iris}
\BIBentryALTinterwordspacing
M.~H. Raju, D.~J. Lohr, and O.~Komogortsev, ``Iris print attack detection using eye movement signals,'' in \emph{2022 Symposium on Eye Tracking Research and Applications}, ser. ETRA '22.\hskip 1em plus 0.5em minus 0.4em\relax New York, NY, USA: Association for Computing Machinery, 2022. [Online]. Available: \url{https://doi.org/10.1145/3517031.3532521}
\BIBentrySTDinterwordspacing

\bibitem{maiorana2015permanence}
E.~Maiorana, D.~La~Rocca, and P.~Campisi, ``On the permanence of eeg signals for biometric recognition,'' \emph{IEEE Transactions on Information Forensics and Security}, vol.~11, no.~1, pp. 163--175, 2015.

\bibitem{labati2013ecg}
R.~D. Labati, R.~Sassi, and F.~Scotti, ``Ecg biometric recognition: Permanence analysis of qrs signals for 24 hours continuous authentication,'' in \emph{2013 IEEE international workshop on information forensics and security (WIFS)}.\hskip 1em plus 0.5em minus 0.4em\relax IEEE, 2013, pp. 31--36.

\bibitem{zhang2021specificity}
Y.~Zhang, M.~Li, H.~Shen, and D.~Hu, ``On the specificity and permanence of electroencephalography functional connectivity,'' \emph{Brain Sciences}, vol.~11, no.~10, p. 1266, 2021.

\bibitem{reliability1}
M.~P. Sampat, G.~Whitman, T.~Stephens, L.~Broemeling, N.~Heger, A.~Bovik, and M.~K. Markey, ``The reliability of measuring physical characteristics of spiculated masses on mammography,'' \emph{The British journal of radiology}, vol.~79, no. special\_issue\_2, pp. S134--S140, 2006.

\bibitem{reliability2}
S.~Noble, D.~Scheinost, and R.~T. Constable, ``A guide to the measurement and interpretation of fmri test-retest reliability,'' \emph{Current opinion in behavioral sciences}, vol.~40, pp. 27--32, 2021.

\bibitem{reliability3}
A.~Caceres, D.~L. Hall, F.~O. Zelaya, S.~C. Williams, and M.~A. Mehta, ``Measuring fmri reliability with the intra-class correlation coefficient,'' \emph{Neuroimage}, vol.~45, no.~3, pp. 758--768, 2009.

\bibitem{brasil2020eye}
\BIBentryALTinterwordspacing
K.~S.~K. Antonio Ricardo Alexandre~Brasil, Jefferson Oliveira~Andrade, ``Eye movements biometrics: A bibliometric analysis from 2004 to 2019,'' \emph{International Journal of Computer Applications}, vol. 176, no.~24, pp. 1--9, May 2020. [Online]. Available: \url{https://ijcaonline.org/archives/volume176/number24/31344-2020920243/}
\BIBentrySTDinterwordspacing

\bibitem{zhang2015survey}
Y.~Zhang and X.~Mou, ``Survey on eye movement based authentication systems,'' in \emph{Computer Vision: CCF Chinese Conference, CCCV 2015, Xi'an, China, September 18-20, 2015, Proceedings, Part I}.\hskip 1em plus 0.5em minus 0.4em\relax Springer, 2015, pp. 144--159.

\bibitem{zhang2016biometrics}
Y.~Zhang and M.~Juhola, ``On biometrics with eye movements,'' \emph{IEEE journal of biomedical and health informatics}, vol.~21, no.~5, pp. 1360--1366, 2016.

\bibitem{Rigas2017}
I.~Rigas and O.~V. Komogortsev, ``Current research in eye movement biometrics: An analysis based on {BioEye} 2015 competition,'' \emph{Image and Vision Computing}, vol.~58, pp. 129--141, 2017.

\bibitem{Andersson2017}
\BIBentryALTinterwordspacing
R.~Andersson, L.~Larsson, K.~Holmqvist, M.~Stridh, and M.~Nystr{\"{o}}m, ``{One algorithm to rule them all? An evaluation and discussion of ten eye movement event-detection algorithms},'' \emph{Behavior Research Methods}, vol.~49, no.~2, pp. 616--637, 2017. [Online]. Available: \url{https://doi.org/10.3758/s13428-016-0738-9}
\BIBentrySTDinterwordspacing

\bibitem{ONH}
M.~Nystr{\"o}m and K.~Holmqvist, ``An adaptive algorithm for fixation, saccade, and glissade detection in eyetracking data,'' \emph{Behavior research methods}, vol.~42, no.~1, pp. 188--204, 2010.

\bibitem{classification1}
J.~Pekkanen and O.~Lappi, ``A new and general approach to signal denoising and eye movement classification based on segmented linear regression,'' \emph{Scientific reports}, vol.~7, no.~1, p. 17726, 2017.

\bibitem{classification2}
A.~H. Dar, A.~S. Wagner, and M.~Hanke, ``Remodnav: robust eye-movement classification for dynamic stimulation,'' \emph{Behavior research methods}, vol.~53, no.~1, pp. 399--414, 2021.

\bibitem{Friedman2018}
\BIBentryALTinterwordspacing
L.~Friedman, I.~Rigas, E.~Abdulin, and O.~V. Komogortsev, ``A novel evaluation of two related and two independent algorithms for eye movement classification during reading,'' \emph{Behavior Research Methods}, vol.~50, no.~4, pp. 1374--1397, 08 2018. [Online]. Available: \url{https://doi.org/10.3758/s13428-018-1050-7}
\BIBentrySTDinterwordspacing

\bibitem{Li2018}
\BIBentryALTinterwordspacing
C.~Li, J.~Xue, C.~Quan, J.~Yue, and C.~Zhang, ``{Biometric recognition via texture features of eye movement trajectories in a visual searching task},'' \emph{PLoS ONE}, vol.~13, no.~4, p. e0194475, apr 2018. [Online]. Available: \url{https://dx.plos.org/10.1371/journal.pone.0194475}
\BIBentrySTDinterwordspacing

\bibitem{rigas2012biometric}
I.~Rigas, G.~Economou, and S.~Fotopoulos, ``Biometric identification based on the eye movements and graph matching techniques,'' \emph{Pattern Recognition Letters}, vol.~33, no.~6, pp. 786--792, 2012.

\bibitem{rigas2016biometric}
I.~Rigas, O.~Komogortsev, and R.~Shadmehr, ``Biometric recognition via eye movements: Saccadic vigor and acceleration cues,'' \emph{ACM Transactions on Applied Perception (TAP)}, vol.~13, no.~2, pp. 1--21, 2016.

\bibitem{holland2013complex}
C.~D. Holland and O.~V. Komogortsev, ``Complex eye movement pattern biometrics: Analyzing fixations and saccades,'' in \emph{2013 International conference on biometrics (ICB)}.\hskip 1em plus 0.5em minus 0.4em\relax IEEE, 2013, pp. 1--8.

\bibitem{george2016score}
A.~George and A.~Routray, ``A score level fusion method for eye movement biometrics,'' \emph{Pattern Recognition Letters}, vol.~82, pp. 207--215, 2016.

\bibitem{Jia2018}
S.~Jia, D.~H. Koh, A.~Seccia, P.~Antonenko, R.~Lamb, A.~Keil, M.~Schneps, and M.~Pomplun, ``{Biometric recognition through eye movements using a recurrent neural network},'' in \emph{Proceedings - 9th IEEE International Conference on Big Knowledge, ICBK 2018}.\hskip 1em plus 0.5em minus 0.4em\relax Institute of Electrical and Electronics Engineers Inc., dec 2018, pp. 57--64.

\bibitem{abdelwahab2022deep}
A.~Abdelwahab and N.~Landwehr, ``Deep distributional sequence embeddings based on a wasserstein loss,'' \emph{Neural Processing Letters}, vol.~54, no.~5, pp. 3749--3769, 2022.

\bibitem{griffith2021gazebase}
H.~Griffith, D.~Lohr, E.~Abdulin, and O.~Komogortsev, ``Gazebase, a large-scale, multi-stimulus, longitudinal eye movement dataset,'' \emph{Scientific Data}, vol.~8, no.~1, p. 184, 2021.

\bibitem{lohr2023gazebasevr}
D.~Lohr, S.~Aziz, L.~Friedman, and O.~V. Komogortsev, ``Gazebasevr, a large-scale, longitudinal, binocular eye-tracking dataset collected in virtual reality,'' \emph{Scientific Data}, vol.~10, no.~1, 2023.

\bibitem{savitzkyGolayM}
A.~Savitzky and M.~J. Golay, ``Smoothing and differentiation of data by simplified least squares procedures.'' \emph{Analytical chemistry}, vol.~36, no.~8, pp. 1627--1639, 1964.

\bibitem{densenet}
G.~Huang, Z.~Liu, L.~Van Der~Maaten, and K.~Q. Weinberger, ``Densely connected convolutional networks,'' in \emph{Proceedings of the IEEE conference on computer vision and pattern recognition}, 2017, pp. 4700--4708.

\bibitem{adam}
D.~P. Kingma and J.~Ba, ``Adam: A method for stochastic optimization,'' \emph{arXiv preprint arXiv:1412.6980}, 2014.

\bibitem{Lr}
L.~N. Smith and N.~Topin, ``Super-convergence: Very fast training of neural networks using large learning rates,'' in \emph{Artificial intelligence and machine learning for multi-domain operations applications}, vol. 11006.\hskip 1em plus 0.5em minus 0.4em\relax SPIE, 2019, pp. 369--386.

\bibitem{Wang2019}
X.~{Wang}, X.~{Han}, W.~{Huang}, D.~{Dong}, and M.~R. {Scott}, ``Multi-similarity loss with general pair weighting for deep metric learning,'' in \emph{2019 IEEE/CVF Conference on Computer Vision and Pattern Recognition (CVPR)}, 2019, pp. 5017--5025.

\bibitem{field2005kendall}
A.~P. Field, ``Kendall's coefficient of concordance,'' \emph{Encyclopedia of Statistics in Behavioral Science}, vol.~2, pp. 1010--11, 2005.

\bibitem{Musgrave2020a}
K.~Musgrave, S.~Belongie, and S.-N. Lim, ``Pytorch metric learning,'' 2020.

\bibitem{luo2016understanding}
W.~Luo, Y.~Li, R.~Urtasun, and R.~Zemel, ``Understanding the effective receptive field in deep convolutional neural networks,'' \emph{Advances in neural information processing systems}, vol.~29, 2016.

\bibitem{prasse}
P.~Prasse, L.~A. J{\"a}ger, S.~Makowski, M.~Feuerpfeil, and T.~Scheffer, ``On the relationship between eye tracking resolution and performance of oculomotoric biometric identification,'' \emph{Procedia Computer Science}, vol. 176, pp. 2088--2097, 2020.

\bibitem{david2021privacy}
B.~David-John, D.~Hosfelt, K.~Butler, and E.~Jain, ``A privacy-preserving approach to streaming eye-tracking data,'' \emph{IEEE Transactions on Visualization and Computer Graphics}, vol.~27, no.~5, pp. 2555--2565, 2021.

\bibitem{lohr2019evaluating}
D.~J. Lohr, L.~Friedman, and O.~V. Komogortsev, ``Evaluating the data quality of eye tracking signals from a virtual reality system: Case study using smi's eye-tracking htc vive,'' \emph{arXiv preprint arXiv:1912.02083}, 2019.

\bibitem{smallsample}
B.~Hanczar, J.~Hua, C.~Sima, J.~Weinstein, M.~Bittner, and E.~R. Dougherty, ``Small-sample precision of roc-related estimates,'' \emph{Bioinformatics}, vol.~26, no.~6, pp. 822--830, 2010.

\end{thebibliography}

\end{document}


\title{Supplementary Material for \textit{Analysis of Embeddings Learned by End-to-End Machine Learning Eye Movement-driven Biometrics Pipeline}}

\maketitle

\section{Experiment Design}

\begin{figure*}[htbp]
\centering
\includegraphics[width=0.9\textwidth]{figures/Ex1.pdf}
\caption{}
\end{figure*}

\begin{figure*}[htbp]
\centering
\includegraphics[width=1\textwidth]{figures/Ex2_4.pdf}
\caption{}
\end{figure*}

\begin{figure*}[htbp]
\centering
\includegraphics[width=1\textwidth]{figures/Ex5.pdf}
\caption{}
\end{figure*}


